\documentclass[10pt,logo,copyright]{nvidiatechreport}
\usepackage{cite}

\hypersetup{colorlinks=true,allcolors=blue}

\newcommand{\agile}{\textsc{agile}}
\newcommand{\isaaclab}{Isaac Lab}

\newcommand{\mujoco}{MuJoCo}
\newcommand{\rslrl}{RSL-RL}

\newcommand{\wandb}{W\&B}
\newcommand{\groot}{GR00T}

\newcommand{\ms}[2]{$#1${\scriptsize$\pm$#2}}

\newcommand{\gone}{Unitree G1}
\newcommand{\tone}{Booster T1}

\newcommand{\ppo}{PPO}
\newcommand{\ltctc}{L2C2}

\newcommand{\policy}{\pi}                       
\newcommand{\valuefn}{V}                        
\newcommand{\obs}{\mathbf{x}}                   
\newcommand{\obspriv}{\mathbf{x}^{\mathrm{p}}} 
\newcommand{\rew}{r}                            
\newcommand{\disc}{\gamma}                      

\newcommand{\interpf}{\alpha}                   
\newcommand{\lactor}{\lambda_{\pi}}             
\newcommand{\lcritic}{\lambda_{V}}              

\newcommand{\rewnorm}{\hat{r}}                  
\newcommand{\ewmastd}{\sigma_r}                 
\newcommand{\decayf}{\beta}                     
\newcommand{\gammafac}{\phi_\disc}              
\newcommand{\retcorr}{c}                        

\newcommand{\fharness}{\mathbf{f}_{\mathrm{h}}} 
\newcommand{\tharness}{\boldsymbol{\tau}_{\mathrm{h}}} 
\newcommand{\qerr}{\mathbf{e}_q}                
\newcommand{\kstiff}{K_p}                       
\newcommand{\kdamp}{K_d}                        
\newcommand{\fscale}{s}                         

\newcommand{\pos}{\mathbf{q}}                   
\newcommand{\postarget}{\mathbf{q}^{*}}         
\newcommand{\emac}{\alpha_{\mathrm{ema}}}       
\newcommand{\velmax}{v_{\max}}                   
\newcommand{\accmax}{a_{\max}}                   

\newcommand{\termsigma}{\sigma}                   

\newcommand{\secref}[1]{Section~\ref{#1}}
\newcommand{\figref}[1]{Figure~\ref{#1}}
\newcommand{\tabref}[1]{Table~\ref{#1}}

\newcommand{\cmark}{\checkmark}
\newcommand{\pmark}{$\sim$}

\title{AGILE: A Comprehensive Workflow \\ for Humanoid Loco-Manipulation Learning}
\author{Huihua Zhao\textsuperscript{*},
Rafael Cathomen\textsuperscript{*},
Lionel Gulich,
Wei Liu,
Efe Arda Ongan,
Michael Lin,
Shalin Jain,
Soha Pouya,
Yan Chang\\
{\small \textsuperscript{*}Equal contribution.}}

\graphicspath{{figures/}}

\begin{abstract}
Recent advances in reinforcement learning (RL) have enabled impressive humanoid behaviors in simulation, yet transferring these results to new robots remains challenging. In many real deployments, the primary bottleneck is no longer simulation throughput or algorithm design, but the absence of systematic infrastructure that links environment verification, training, evaluation, and deployment in a coherent loop. 

To address this gap, we present AGILE, an end-to-end workflow for humanoid RL that standardizes the policy-development lifecycle to mitigate common sim-to-real failure modes. AGILE comprises four stages: (1) interactive environment verification, (2) reproducible training, (3) unified evaluation, and (4) descriptor-driven deployment via robot/task configuration descriptors. For evaluation stage, AGILE supports both scenario-based tests and randomized rollouts under a shared suite of motion-quality diagnostics, enabling automated regression testing and principled robustness assessment. AGILE also incorporates a set of training stabilizations and algorithmic enhancements in training stage to improve optimization stability and sim-to-real transfer. 

With this pipeline in place, we validate AGILE across five representative humanoid skills spanning locomotion, recovery, motion imitation, and loco-manipulation on two hardware platforms (Unitree G1 and Booster T1), achieving consistent sim-to-real transfer. Overall, AGILE shows that a standardized, end-to-end workflow can substantially improve the reliability and reproducibility of humanoid RL development.

Code: \url{https://github.com/nvidia-isaac/WBC-AGILE}
\end{abstract}

\begin{document}
\maketitle
\abscontent

\section{Introduction}
\label{sec:introduction}
Reinforcement learning has enabled increasingly capable humanoid locomotion and manipulation policies~\cite{rudin2022learning, cheng2024expressive,
radosavovic2024humanoid, he2024learning, chen2025gmtgeneralmotiontracking, luo2025sonicsupersizingmotiontracking}, yet translating these results to new robots and tasks remains fragile and labor-intensive. In practice, failures rarely stem from insufficient simulation throughput or algorithmic novelty, but from the absence of structured infrastructure connecting environment verification, scalable training, systematic evaluation, and deployment.

\textbf{The Workflow Gap:} Humanoid RL development is often built on fragmented and ad hoc workflows. Basic environment issues, such as reversed joint axes or incorrect reward terms, are frequently discovered only after costly training runs. Policy evaluation is also commonly performed through stochastic rollouts, which measure average task performance under randomized commands but can make it difficult to diagnose hardware-critical behaviors such as joint limit violations or high-frequency actuation. As a result, the lifecycle of humanoid RL development, from environment verification to deployment, remains poorly structured and hard to reproduce.

 \textbf{The Transfer Gap:} Exporting a learned policy for external validation or hardware deployment is a notoriously fragile process. Without a standardized I/O contract, researchers must manually resolve joint order mismatches, reconstruct observation history buffers, and align action scaling. This ad hoc translation introduces silent bugs and prevents the use of a unified evaluation pipeline across secondary simulators (like MuJoCo), forcing researchers to risk physical deployment without rigorous, quantitative pre-validation.

\begin{figure*}[t]
    \centering
    \includegraphics[width=\linewidth]{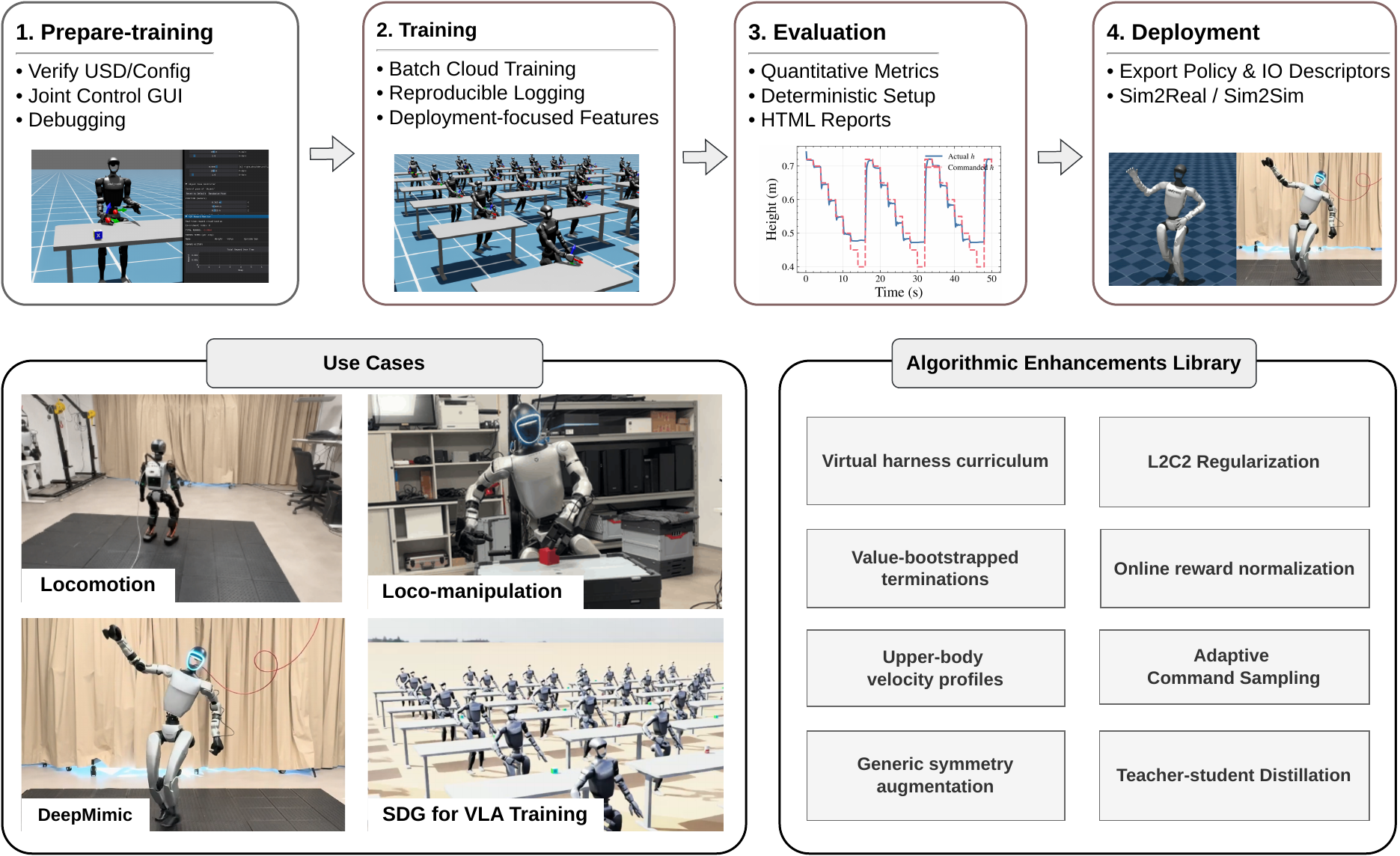}
    \caption{Overview of \agile{} learning workflow. The workflow covers prepare-training, batch cloud training with reproducible logging and deployment-oriented features, evaluation using quantitative motion metrics and automated HTML reports, and deployment by exporting the learned policy and I/O descriptors for Sim2Sim and Sim2Real transfer. Example applications include locomotion, loco-manipulation, DeepMimic-style imitation, and synthetic data generation (SDG) for VLA training, supported by an algorithmic enhancements library (e.g., curricula, regularization, adaptive sampling, reward normalization, symmetry augmentation, and distillation).}
    \label{fig:pipeline}
\end{figure*}

To resolve these bottlenecks, we present AGILE (A Generic Isaac-Lab based Engine), an open-source workflow built on Isaac Lab and RSL-RL that covers the full path from a new robot to a deployed policy. AGILE bridges the workflow and transfer gaps through a four-stage pipeline (\figref{fig:pipeline}):
\begin{enumerate}
  \item \textbf{Prepare:} Interactive debug GUIs (for joint control, object manipulation, and reward visualization) allow researchers to catch robot model and MDP misconfigurations in minutes before committing GPU hours.
  \item \textbf{Train:} A scalable, reproducible training environment featuring automated hyperparameter sweeps, experiment tracking, and a suite of independently toggleable algorithmic enhancements.
  \item \textbf{Evaluate:} A unified evaluation pipeline combining deterministic scenario-based tests and stochastic rollouts. Parallel environments receive scripted or randomized commands to evaluate deployment-critical motion metrics such as joint jerk and limit violations.
  \item \textbf{Deploy:} Trained policies are auto-exported alongside self-contained YAML I/O descriptors that resolve joint ordering and action scaling. This powers a unified inference pipeline for both quantitative sim-to-sim validation in MuJoCo and real-world hardware deployment.
\end{enumerate}

Beyond infrastructure, \agile{} also packages a suite of training enhancements for sim-to-real transfer (\ltctc{}~\cite{kobayashi2022l2c2}, reward normalization, value-bootstrapped terminations, symmetry augmentation~\cite{mittal2024symmetry}, virtual harness), each validated through thorough ablations. To further showcase \agile{}'s modularity, we present a decoupled whole-body control application~\cite{ben2025homiehumanoidlocomanipulationisomorphic, li2025amoadaptivemotionoptimization}, in which a frozen locomotion policy serves as a lower-body API while an independent upper-body expert collects demonstration data for VLA fine-tuning~\cite{lu2025mobiletelevisionpredictivemotionpriors}.

Conceptually, AGILE reframes humanoid reinforcement learning as a structured engineering lifecycle rather than a collection of loosely connected scripts. By formalizing interfaces across verification, training, evaluation, and deployment, the framework enables deterministic regression testing, deployment-oriented motion diagnostics, and descriptor-consistent policy export prior to hardware trials. This shifts humanoid RL development from empirical trial-and-error toward repeatable, quantitatively validated engineering.

We validate the complete workflow across five tasks on the \gone{} and \tone{}: velocity tracking, height-controlled locomotion~\cite{ben2025homiehumanoidlocomanipulationisomorphic}, stand-up~\cite{huang2025learninghumanoidstandingupcontrol, he2025learninggettinguppoliciesrealworld}, motion imitation~\cite{ji2025exbody2advancedexpressivehumanoid, sun2025robotdancingresidualactionreinforcementlearning}, and loco-manipulation with VLA~\cite{xu2024humanvlavisionlanguagedirectedobject}. Our contributions are:
\begin{itemize}
  \item \textbf{A structured lifecycle for humanoid RL}, integrating environment verification, training, evaluation, and descriptor-driven deployment into a unified workflow.
  \item \textbf{A unified evaluation framework} combining deterministic scenario tests and stochastic rollouts with per-joint motion-quality metrics (jerk, limit violations) for quantitatively regression test and deployment-oriented policy validation.
  \item \textbf{Validation across five tasks and two platforms}, with sim-to-real transfer for locomotion, recovery, imitation, and loco-manipulation, released as open-source with pre-trained checkpoints.
\end{itemize}

\section{Related Work}
\label{sec:related_work}

We categorize related work into simulation platforms, humanoid learning frameworks, algorithmic techniques for sim-to-real transfer, and evaluation methodologies.

\subsection{GPU-Accelerated Simulation Primitives}
GPU-based simulators have significantly accelerated reinforcement learning for robotics. Isaac Gym~\cite{makoviychuk2021isaac} introduced tensor-based simulation pipelines that eliminate CPU--GPU bottlenecks, while Isaac Lab~\cite{nvidia2025isaaclabgpuacceleratedsimulation} extends this approach with a modular manager-based architecture and USD-based scene configuration. Parallel efforts such as MuJoCo Playground~\cite{zakka2025mujoco} bring GPU acceleration to the MuJoCo physics engine. While these platforms provide powerful simulation primitives, they primarily focus on simulation performance and environment modeling rather than the workflow surrounding debugging, evaluation, and deployment.

\subsection{Humanoid Learning Frameworks}
Several recent frameworks aim to scale humanoid learning pipelines. Holosoma focuses on large-scale infrastructure and fast off-policy learning for humanoid locomotion~\cite{amazon-holosoam, seo2025fasttd3}. HumanoidVerse emphasizes cross-simulator compatibility through abstraction layers across multiple physics engines~\cite{humanoidverse2025}, ProtoMotions provides an Isaac Lab-native framework for motion tracking and humanoid control~\cite{ProtoMotions}, and RoboVerse provides unified interfaces for scalable robot learning across tasks and embodiments~\cite{geng2025roboverse}. These frameworks primarily address training scalability and simulator interoperability. In contrast, \agile{} focuses on the broader development lifecycle of humanoid RL policies, including environment verification, deterministic evaluation, and deployment. \tabref{tab:framework_comparison} summarizes key differences.

\begin{table}[t]
\centering
\caption{Feature comparison of humanoid RL frameworks. \cmark = supported, \pmark = partial, -- = not supported.}
\label{tab:framework_comparison}
\small
\setlength{\tabcolsep}{3pt}
\begin{tabular}{@{}lcccc@{}}
\toprule
Feature & \agile{} & Holosoma & H.Verse & Proto. \\
\midrule
Env.\ debugging       & \cmark & --     & --     & --     \\
Algo.\ enhancements      & \cmark & \pmark & --     & \pmark \\
Determ.\ evaluation      & \cmark & \pmark & --     & \cmark \\
Sim-to-sim pipeline    & \cmark & \cmark    & \cmark & \cmark     \\
Descriptor-driven export & \cmark & --     & --     & --     \\
Sim-to-real validation    & \cmark & \cmark    & \cmark & \cmark     \\
Multi-sim backend support    & -- & --     & \cmark & \cmark     \\
\bottomrule
\end{tabular}
\end{table}

\subsection{Algorithmic Techniques for Sim-to-Real Transfer}
A number of approaches improve policy robustness for real-world deployment. CAPS encourages smoother control policies through action regularization~\cite{mysore2021caps}, while L2C2 enforces local Lipschitz continuity to improve stability under observation perturbations~\cite{kobayashi2022l2c2}. Other approaches such as ASAP use data from real world deployments to learn a residual action policy that enables simulation dynamics to better align with real world dynamics~\cite{song2025asap}. While these techniques improve policy stability, they are typically applied independently. AGILE instead integrates such stabilization methods within a unified training and evaluation workflow.

\subsection{Evaluation and Benchmarking}
Earlier benchmark suites such as OpenAI Gym \cite{brockman2016gym} and the DeepMind Control Suite \cite{tassa2018dmcontrol} established standardized task evaluation in simulation, motivating similar rigor for humanoid systems. HumanoidBench provides simulation benchmarks for locomotion and manipulation tasks~\cite{gu2024humanoidbench}, while RoboGauge proposes a predictive assessment suite for quantifying sim-to-real transferability in quadrupedal locomotion~\cite{wu2026robogauge}. Related efforts from the IEEE Humanoid Study Group seek standardized metrics for stability and safety in humanoid systems. AGILE builds on this direction by incorporating a unified evaluation framework into the RL development workflow, supporting deterministic scenario tests and stochastic rollouts with motion-quality diagnostics.

\section{System Design}
\label{sec:system_design}
\subsection{Overview}
\label{sec:sysdesign-overview}

\agile{} is a comprehensive workflow layer built on top of \isaaclab{} (providing parallel GPU simulation and MDP primitives) and \rslrl{} (providing RL algorithms). It adds a four-stage pipeline (\figref{fig:pipeline}) that wraps the training loop with tooling for verification, reproducibility, evaluation, and deployment.

The workflow follows a configuration-driven, flat architecture: every task is a self-contained file specifying the scene, observations, actions, rewards, terminations, and curriculum. Because every MDP parameter can be modified directly via configuration, researchers can rapidly prototype, sweep parameters, and deploy policies without structural code changes.

\subsection{Training Preparation}
\label{sec:prepare}

Misconfigurations such as incorrect joint directions, collision geometries, or reward terms can waste days of GPU time. \agile{} provides three composable GUI plugins, built on Isaac Lab’s manager terms, that attach to any environment for interactive pre-training validation:

\textbf{Joint Position GUI.} Per-joint slider control with real-time torque readout; an optional symmetry mode displays mirrored robots side by side to spot sign errors in roll/yaw axes.

\textbf{Object Manipulation GUI.} 6-DOF object positioning with live contact-sensor visualization for verifying that manipulation-based rewards activate correctly.

\textbf{Reward Visualizer.} Per-term reward overlay showing each component’s weight and contribution while users manipulate the scene, confirming reward behavior without running a training loop.

\begin{figure}[t]
    \centering
    \includegraphics[width=0.75\linewidth]{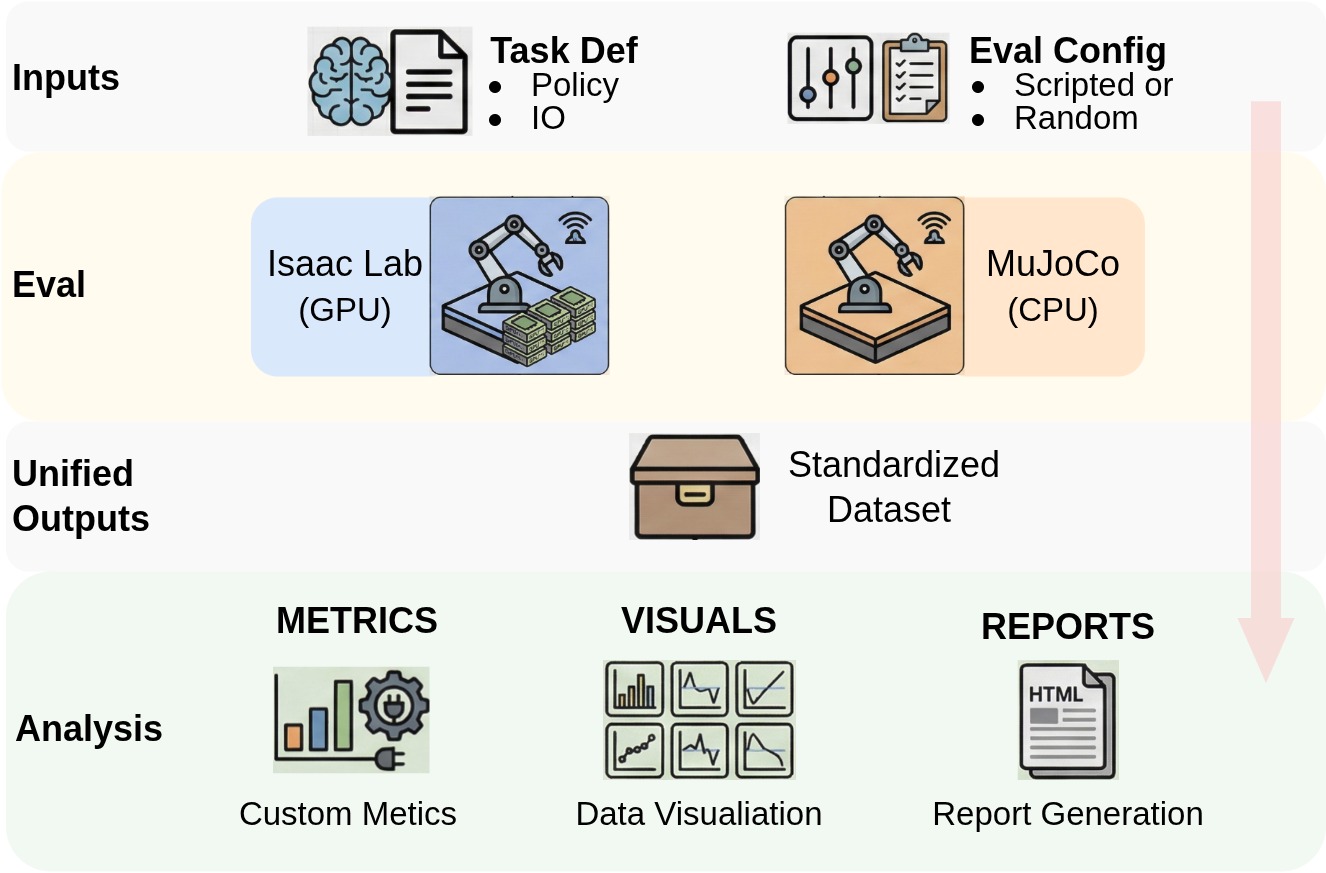}
    \caption{Deterministic Evaluation Pipeline. A unified framework for assessing humanoid policies across \isaaclab{} (GPU) and \mujoco{} (CPU) backends.}
    \label{fig:eval-pipeline}
\end{figure}

\subsection{Training}
\label{sec:training}

A unified entry point manages training, evaluation, and parameter sweeps, supporting both local execution and cloud deployment.

\subsubsection{Training Infrastructure}
To ensure reproducibility, \agile{} records a lightweight git snapshot (commit hash, branch, and uncommitted diffs) together with YAML configuration dumps for every run. Combined with Docker-based orchestration and \wandb{} logging, experiments are exactly reproducible.

AGILE also supports structured hyperparameter sweeps through \emph{scaled-dict parameters}. Instead of independently sweeping each entry of structured parameter groups (e.g., leg PD gains), scaled-dict allowed for joint-parameter sweeping by scaling a single scaling parameter that is applied to the entire dictionary, preserving relative structure while collapsing the search into a one-dimensional variable. Because AGILE builds on Isaac Lab’s manager architecture, any MDP parameter, not only RL hyperparameters, can participate in these sweeps.

\subsubsection{Algorithmic Toolbox} AGILE integrates several commonly used stabilization techniques as independently toggleable modules within the training pipeline.

\textbf{\ltctc{} Regularization.}
Given consecutive observations $(\obs_t, \obs_{t+1})$, we form an
interpolated input
$\tilde{\obs} = \obs_t + \interpf\,(\obs_{t+1} - \obs_t)$ with
$\interpf \sim \mathcal{U}(0,1)$ and penalize the output change:%
\begin{equation}\label{eq:l2c2_loss}
  \mathcal{L}
    = \lactor  \left\lVert \policy(\tilde{\obs}) - \policy(\obs_t) \right\rVert^2
    + \lcritic \left\lVert \valuefn(\tilde{\obspriv}) - \valuefn(\obspriv_t) \right\rVert^2,
\end{equation}
enforcing local Lipschitz continuity~\cite{kobayashi2022l2c2} for smooth,
hardware-safe actions, also used
in~\cite{huang2025learninghumanoidstandingupcontrol}.

\textbf{Online Reward Normalization.}
Rewards are normalized~\cite{andrychowicz2021matters} by a running standard deviation,
\begin{equation}
\rewnorm_t =
\frac{\rew_t}{\ewmastd \cdot \gammafac \cdot \retcorr + \epsilon},
\end{equation}
where $\epsilon = 10^{-2}$ prevents division by zero, $\ewmastd$ is an EMA standard deviation across the environment batch, $\gammafac = 1/\sqrt{1-\disc^2}$ accounts for discounted return variance, and $\retcorr$ is a return-scale correction updated as $\retcorr \leftarrow \decayf\,\retcorr + (1-\decayf)\,\sigma_G \cdot \retcorr$, where $\sigma_G$ is the GAE return standard deviation. Because rewards are divided by $\retcorr$, the product $\sigma_G \cdot \retcorr$ is invariant to the current normalization, making training largely invariant to reward magnitude changes during curriculum.

\textbf{Value-Bootstrapped Terminations.}
Standard GAE bootstraps the value to zero at termination. The conventional remedy---a sparse penalty $p$---is fragile: $p$ must satisfy $p < V(\obs_T)$ for all terminal states, otherwise the agent prefers dying over continuing when expected returns are negative. We instead modify the terminal reward as
\begin{equation}\label{eq:term_bootstrap_main}
  \rewnorm_T \leftarrow \rewnorm_T + \disc\,V(\obs_T) +
  \begin{cases}
    -\termsigma & \text{bad (e.g.\ falling)} \\
    +\termsigma & \text{good (e.g.\ reaching goal)} \\
    \phantom{+}0 & \text{neutral (e.g.\ timeout)}
  \end{cases}
\end{equation}

The $\disc V(\obs_T)$ term makes termination value-neutral (as if the episode continued), while $\termsigma>0$ shifts the outcome to be strictly worse or better than continuing. Because $\termsigma$ operates after reward normalization, it remains scale-invariant ($\termsigma{=}5$ for all tasks). This is related to potential-based shaping~\cite{ng1999policy}, applied only at terminal states. The modified Bellman operator is a $\disc$-contraction; with $\disc{=}0.99$, a bad termination offset of $\termsigma{=}5$ amplifies to an effective shift of $500$ in value space.

\textbf{Virtual Harness.}\label{sec:harness_force}
Much like a physical harness that supports a person learning to walk, external PD forces applied to the root body stabilize the robot during early training, preventing immediate collapse before the policy can discover useful behaviors:
$\tharness = \kstiff\,\qerr - \kdamp\,\boldsymbol{\omega}$,\;
$\fharness = \kstiff\,(h^{*} - h) - \kdamp\,\dot{h}$,
where $\kstiff$/$\kdamp$ are proportional/derivative gains, $\qerr$ is
the orientation error to upright, $\boldsymbol{\omega}$ the angular
velocity, and $h^{*}$/$h$ the desired/current root height. A curriculum scale
$\fscale \in [0,1]$ multiplies all gains and limits; supported schedules
are linear decay ($\fscale = 1 - t/T$), exponential decay
($\fscale = e^{(t/T)\ln 0.01}$), and an adaptive variant that decreases
$\fscale$ only when the standing ratio exceeds a threshold.

\begin{figure}[t]
    \centering
    \includegraphics[width=0.95\linewidth]{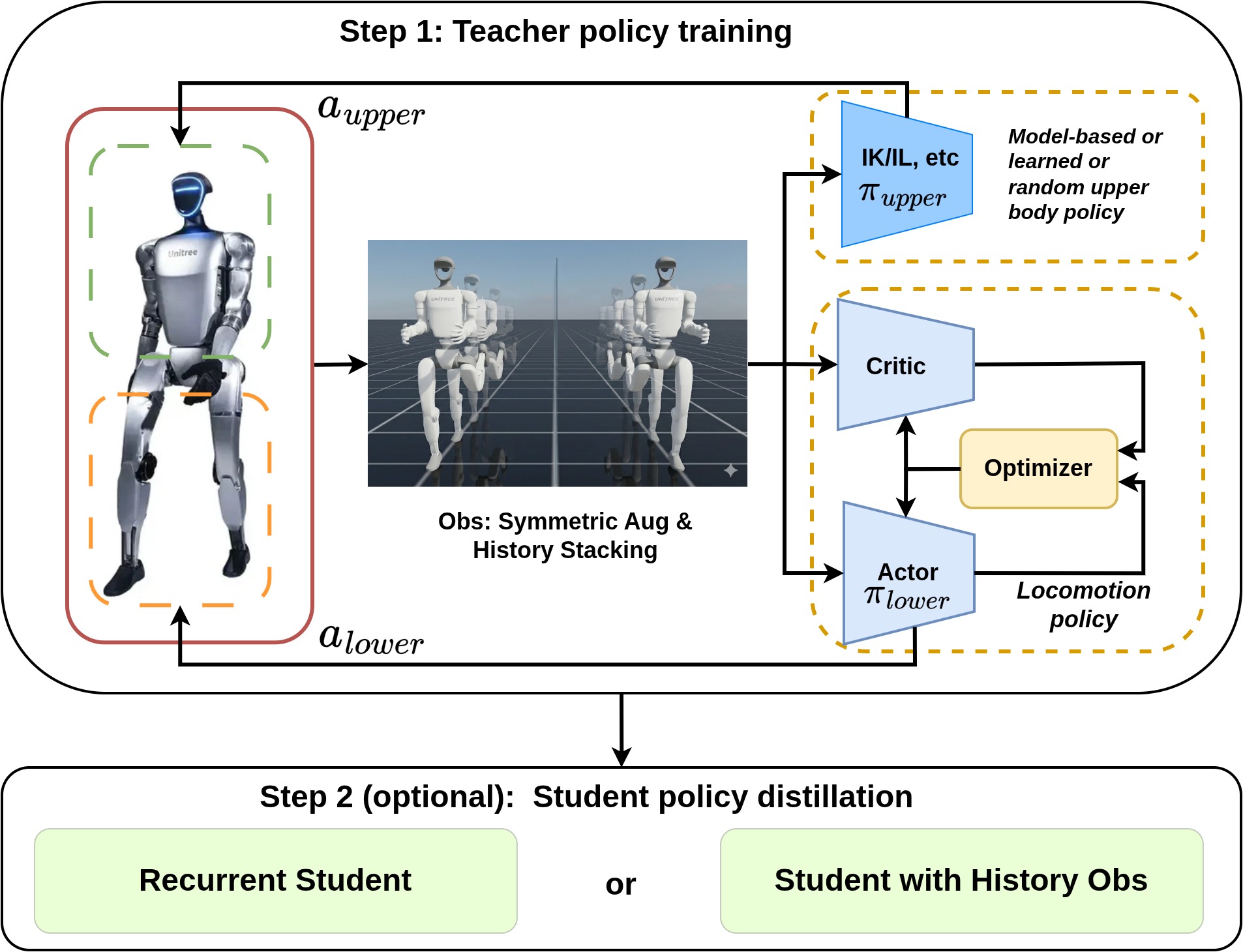}
    \caption{Decoupled whole-body control: upper- and lower-body policies are trained separately. This design allows greater flexibility to meet different application requirements. For example, an IK policy can be used for high-accuracy tasks while a VLA upper-body policy enables autonomous execution with language input.}
    \label{fig:separated_control}
\end{figure}

\textbf{Velocity Profile.} \label{sec:velocity_profile}
When randomizing upper-body joint targets during locomotion training, abrupt
position jumps can destabilize the lower-body policy. \agile{} provides
pluggable velocity profiles that interpolate between the current position
$\pos_t$ and a sampled target $\postarget$:
\begin{itemize}
  \item \textbf{EMA.} Exponential smoothing:
        $\pos_{t+1} = \emac\,\postarget + (1-\emac)\,\pos_t$, with
        $\emac$ sampled uniformly per trajectory. Smooth, zero-overshoot.
  \item \textbf{Trapezoidal.} Three-phase motion with bounded
        $\accmax$ and $\velmax$: accelerate, cruise, decelerate. Joints can be
        synchronized to finish simultaneously. Physically realistic.
  \item \textbf{Linear.} Constant-velocity interpolation. Simplest;
        suitable for non-critical joints.
\end{itemize}
All profiles support per-joint position and velocity limits.

\textbf{Symmetry Augmentation.} Observations and actions are mirrored to encourage symmetric locomotion and effectively double training data~\cite{mittal2024symmetry}. The mapping is configuration-driven rather than index-based, enabling adaptation to new observation spaces and robot morphologies.

\begin{table*}[t]
\centering
\caption{MDP overview and training time (on a single L40) per task.}
\label{tab:mdp_overview}
\small
\begin{tabular}{@{}lccccccc@{}}
\toprule
Task & Robot & Actions & Observation (pol/crit) & Reward dim & Episode & Terrain & Training \\
\midrule
Locomotion       & G1 & Leg pos (12) & noisy / + scans & 15 & 30\,s & rough  & 10h\\
Locomotion       & T1 & Leg pos (10) & noisy / + vel,ht & 20 & 30\,s & rough & 10h \\
Loco + Height    & G1 & Leg pos (12) & noisy / + scans & 22 & 30\,s & rough & 10h \\
Stand-up         & G1 & All joints (rel) & noisy / + forces & 18 & 15\,s & stand-up & 25h \\
Stand-up         & T1 & All joints (rel) & noisy / + forces & 18 & 15\,s & stand-up & 15h \\
Motion Imitation & G1 & All joints (rel) & noisy / + vel & 9 & 8\,s & flat & 6h \\
Pick \& Place    & G1 & Upper body   & tracking / --    & 10 & 25\,s & flat & 10h \\
\bottomrule
\end{tabular}
\end{table*}

\textbf{Additional Modules.}
These include adaptive command
sampling (biasing toward low-speed balancing), upper-body velocity
profiles (EMA, trapezoidal, linear interpolation), state caching for efficient resets (a one-time rollout collects diverse initial states, e.g.\ fallen poses, avoiding repeated drop simulations during training), and teacher-student distillation following \rslrl{} with symmetry mirror loss.

Transferring policies to real hardware requires both \emph{smooth} and \emph{robust} control. AGILE encourages smoothness through action regularization penalties (norm, rate, acceleration) and \ltctc{}, reducing high-frequency actions that real actuators cannot reliably track. Robustness is improved through domain randomization—randomizing dynamics, mass properties, contact parameters, and actuator delays—as established in sim-to-real transfer literature \cite{tobin2017domain, peng2018sim2real}, together with external perturbations and sensor noise injection. These techniques are not new learning algorithms; instead, AGILE provides a unified implementation and empirical characterization of commonly used stabilization methods within a reproducible humanoid RL workflow.

\subsection{Evaluation}
\label{sec:evaluation}
Comparing policies solely with stochastic rollouts can mask failure modes that become visible only under controlled conditions. \agile{} therefore complements stochastic rollouts with deterministic scenario-driven testing, where parallel environments receive identical scripted command sequences (e.g., velocity sweeps or height ramps). The resulting evaluation pipeline combines stochastic rollouts, deterministic scenario tests, and motion-quality diagnostics within a unified workflow. Crucially, this pipeline operates seamlessly in both Isaac Lab and \mujoco{}, allowing the same evaluation scenarios and metrics to be applied during sim-to-sim validation (\figref{fig:eval-pipeline}).

Beyond aggregate task tracking, the pipeline analyzes deployment-critical per-joint motion-quality metrics such as RMS acceleration, jerk, and joint-limit violations. Deterministic scenarios provide reproducible, lower-variance benchmarks for regression testing, while stochastic rollouts evaluate robustness under randomized command distributions. All evaluation results are exported as standalone interactive HTML reports, enabling rapid diagnosis of behaviors that may threaten hardware safety prior to deployment.


\subsection{Deployment}
\label{sec:deployment}

\agile{} builds on \isaaclab{}'s I/O descriptor system, which exports trained policies to TorchScript alongside auto-generated YAML configurations capturing the full I/O contract (joint names, observation ordering, history buffers, action scaling). On top of this, \agile{} provides export tooling and a complete sim-to-sim validation pipeline in \mujoco{} that reads the descriptors to automatically reconstruct observation assembly, action mapping, and history buffers for inference. Hardware-specific driver integrations reuse the same I/O contract, so the core inference logic remains identical; only the state provider changes. The descriptor also facilitates translating the inference stack from Python to C++ for real-time execution on hardware.

\section{Case Studies \& Results}
\label{sec:case_studies}
\begin{figure}[t]
    \centering
    \includegraphics[width=\linewidth]{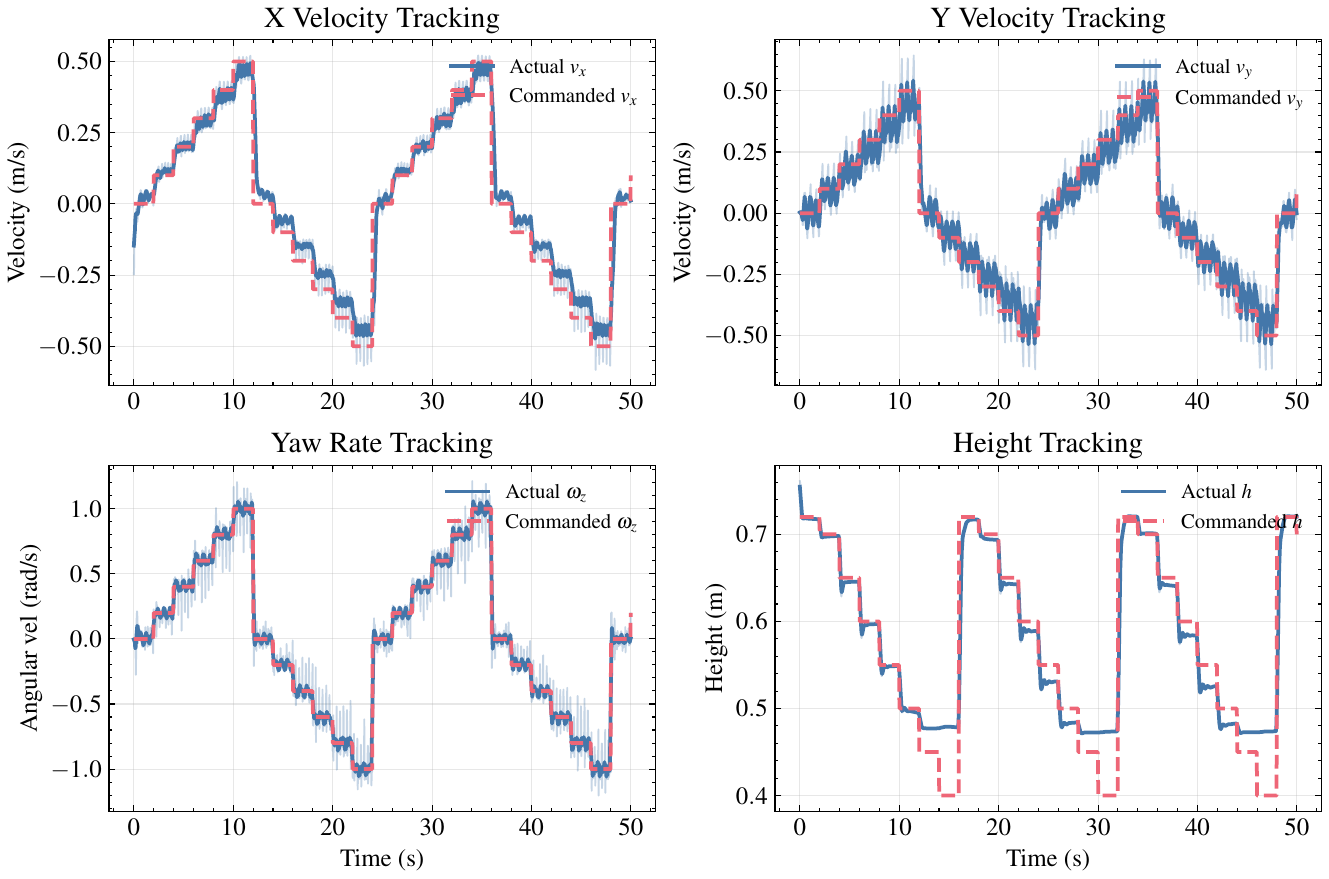}
    \caption{Velocity and height command tracking in MuJoCo. Each subplot shows a sweep of commanded values (dashed) versus actual robot response (solid) for forward velocity, lateral velocity, yaw rate, and standing height. The solid line is smoothed with the actual showing as the lighter shadow for easier visualization.}
    \label{fig:vel_height_tracking}
\end{figure}

We validate \agile{} across five representative tasks using the \gone{} and \tone{} platforms. \tabref{tab:mdp_overview} provides a comprehensive summary of the MDP configurations and the estimated training hours required for each task on a single L40. Although the reward signal often reaches convergence relatively early, we typically extend training runs until $20$k steps reached to ensure optimal policy performance and stability.

\subsection{Velocity Tracking on \gone{} and \tone{}}
\label{sec:locomotion}

The baseline locomotion policy tracks commanded $(v_x, v_y, \omega_z)$ using leg-joint position offsets (\figref{fig:robot_motion}a,b).
To encourage robustness, upper-body joints receive random targets via trapezoidal velocity profiles. This task demonstrates \agile{}'s generic configuration architecture. Both the \gone{} and \tone{} use the exact same MDP template and training pipeline. We utilize a preparation module to ensure the defined symmetry terms are correct, allowing us to safely apply \agile{}'s symmetry augmentation to double the effective training data and enforce symmetric gaits. Additionally, the virtual harness is utilized for early stabilization, preventing initial episode collapse. All locomotion tasks train on procedurally generated rough terrain with an adaptive difficulty curriculum~\cite{rudin2022learning}.

\subsection{Height-Controlled Locomotion on \gone{}}
\label{sec:locomotion_height}

\begin{table}[t]
  \centering
  \caption{%
  Sim-to-Sim evaluation of tracking on \gone{} velocity+height control.
  Left: deterministic sweep (50\,s) vs.\ random commands at
  increasing durations (2\,s resample interval) over 10\,runs.
  Right: teacher vs.\ distilled student policies
  (deterministic sweep, 50\,s).}
  \label{tab:tracking_rmse}
  \resizebox{\columnwidth}{!}{%
  \begin{tabular}{@{}l c ccc c cc@{}}
\toprule
    & \multicolumn{4}{c}{Teacher (Privileged)} & \multicolumn{2}{c}{Student (Distilled)} \\
    \cmidrule(lr){2-5} \cmidrule(lr){6-7}
    & Det. & \multicolumn{3}{c}{Random} & \multicolumn{2}{c}{Deterministic} \\
    \cmidrule(lr){2-2} \cmidrule(lr){3-5} \cmidrule(lr){6-7}
    Duration & 50s & 50s & 200s & 500s & RNN (50s) & Hist. (50s) \\
    \midrule
    $v_x$ (m/s)        & \textbf{0.070} & \ms{0.142}{0.019} & \ms{0.136}{0.010} & \ms{0.136}{0.003} & 0.116 & 0.097 \\
    $v_y$ (m/s)        & \textbf{0.083} & \ms{0.118}{0.013} & \ms{0.110}{0.006} & \ms{0.110}{0.002} & 0.110 & 0.087 \\
    $\omega_z$ (rad/s) & \textbf{0.074} & \ms{0.116}{0.013} & \ms{0.113}{0.007} & \ms{0.112}{0.003} & 0.117 & 0.079 \\
    $h$ (m)            & \textbf{0.035} & \ms{0.046}{0.007} & \ms{0.039}{0.003} & \ms{0.036}{0.001} & 0.037 & 0.037 \\
    \bottomrule
  \end{tabular}%
  }
\end{table}

This task extends velocity tracking by introducing a commanded pelvis height, requiring the robot to maintain stable locomotion across postures ranging from a deep crouch to a full stance (\figref{fig:robot_motion}a).

\textbf{Decoupled WBC architecture.} A critical design choice is ``separated body control.'' The RL policy controls the leg joints exclusively, while the waist and upper-body joints are randomized using a trapezoidal velocity profile during training. This deliberately reserves the upper degrees of freedom, enabling an independent Inverse Kinematics (IK) or Vision-Language-Action (VLA) controller to take full ownership of the torso and arms at deployment (\figref{fig:separated_control}).

This task also serves as a case study for \agile{}'s unified evaluation pipeline (\secref{sec:evaluation}), which combines deterministic scenario tests, stochastic rollouts, and shared motion-quality diagnostics. Policies are first evaluated in IsaacLab; we observed that consistent violations of joint limits reliably preclude successful sim-to-sim transfer; policies are thus fine-tuned using these feedback signals. The pipeline then transitions to MuJoCo, employing scripted height-ramp and velocity-sweep scenarios to provide reproducible, low-variance benchmarks for controlled comparison, while stochastic command sampling assesses robustness. As shown in \tabref{tab:tracking_rmse}, deterministic scenarios produce consistent tracking metrics over short horizons, whereas stochastic rollouts have bigger variance (std is computed with $10$ runs) and require substantially longer durations to converge. This unified evaluation enables consistent quantitative comparisons between the privileged teacher policy and distilled LSTM and history-MLP student architectures (\figref{fig:vel_height_tracking} and \tabref{tab:tracking_rmse}), demonstrating that the evaluation pipeline generalizes across different policy structures.

\subsection{Stand-Up on \tone{} \& \gone{}}
\label{sec:standup}


Fall recovery is a complex whole-body control problem where the policy must coordinate all joints to transition from a random fallen configuration to a stable standing pose~\cite{huang2025learninghumanoidstandingupcontrol}. Both policies successfully transfer to real hardware (\figref{fig:robot_motion}c,d). To handle the massive scale differences between fine-grained postural rewards and the large sparse bonus for standing, this task relies on online reward normalization. A pre-collected dataset of diverse fallen poses (via \agile{}'s state caching) provides varied initial configurations without wasting training compute on repeated drops.

\subsection{Motion Imitation on \gone{}}
\label{sec:dancing}

We formulate a BeyondMimic-style~\cite{liao2025beyondmimicmotiontrackingversatile} motion imitation task to demonstrate that \agile{} generalizes beyond standard command-tracking. The policy tracks an $8s$ dancing sequence by sampling harder segments more frequently. Because the actor uses only hardware-available observations (no privileged state), the policy deploys directly to the real \gone{} hardware without distillation (\figref{fig:robot_motion}e). The default BeyondMimic setup with our motion reference data failed to transfer even to simulation. Therefore, we applied additional domain randomization and L2C2 regularization to support sim-to-real transfer and suppress high-frequency actuator oscillations during real-world deployment.

\subsection{Loco-Manipulation \& VLA Fine-Tuning}
\label{sec:loco_api}

\begin{figure}[t]
    \centering
    \includegraphics[width=0.95\linewidth]{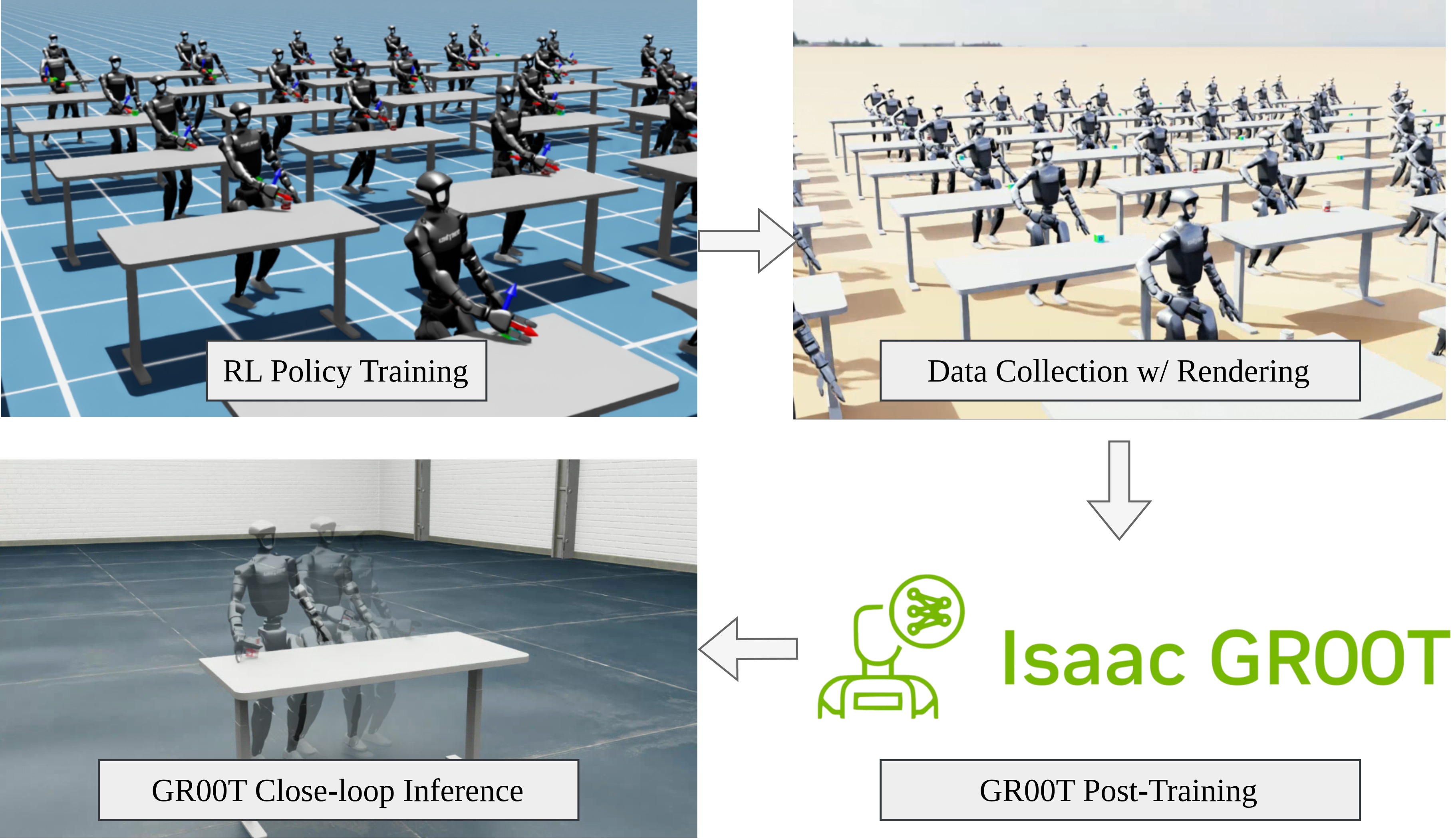}
    \caption{\groot{} post-training pipeline.
    (1)~RL expert trained with physics randomization.
    (2)~Demonstration collection via tiled parallel rendering with scene randomization.
    (3)~VLA fine-tuning on the collected dataset.
    (4)~Closed-loop evaluation in simulation.}
    \label{fig:groot_fine_tune}
\end{figure}

\agile{}'s modular architecture freezes the locomotion policy from
\secref{sec:locomotion_height} as a lower-body controller while an
upper-body module is developed independently. We demonstrate this with a
pick-and-place task on \gone{}, followed by VLA fine-tuning
(\figref{fig:groot_fine_tune}). 

\textbf{RL expert.}
An RL policy controls the right arm and waist only, guided by reference
trajectories and privileged simulation state (object pose, hand--object
distance). Restricting the action space to the upper body reduces training to
20k iterations. Object identity is varied across environments to promote
visual diversity in the resulting demonstrations.

\textbf{VLA data collection and fine-tuning.}
The RL expert generates 100 successful trajectories via tiled parallel
simulation under physics and visual domain randomization, producing paired
RGB observations, proprioception, and actions without human teleoperation.
A \groot{} N1.5 VLA model is fine-tuned on this dataset, replacing privileged
inputs with RGB and language task descriptions. At deployment, the VLA
predicts upper-body targets while the frozen locomotion policy maintains
lower-body stability.

In closed-loop simulation, the post-trained VLA reliably picks up the target object and places it to the side, achieving a 90\% success rate across 100 test cases with randomly sampled initial robot states. These results validate effective transfer from privileged RL to perception-driven control.
\section{Discussion \& Limitations}
\label{sec:discussion}

\subsection{Ablation Studies}
\label{sec:ablations}

We ablate five key enhancements independently.

\subsubsection{Reward Normalizer}
\label{sec:ablation_reward_norm}

Online reward normalization makes training magnitude-agnostic, which is useful when reward scales shift during curriculum progression or when porting configurations across robots.
\figref{fig:combined_ablations}(a) validates this on \gone{} velocity tracking: at the original scale (1$\times$), the normalizer provides a modest improvement, while at 100$\times$ scale it recovers near-original performance.

\subsubsection{\ltctc{}}
\label{sec:ablation_l2c2}

\begin{figure}[t]
    \centering
    \includegraphics[width=1.0\linewidth]{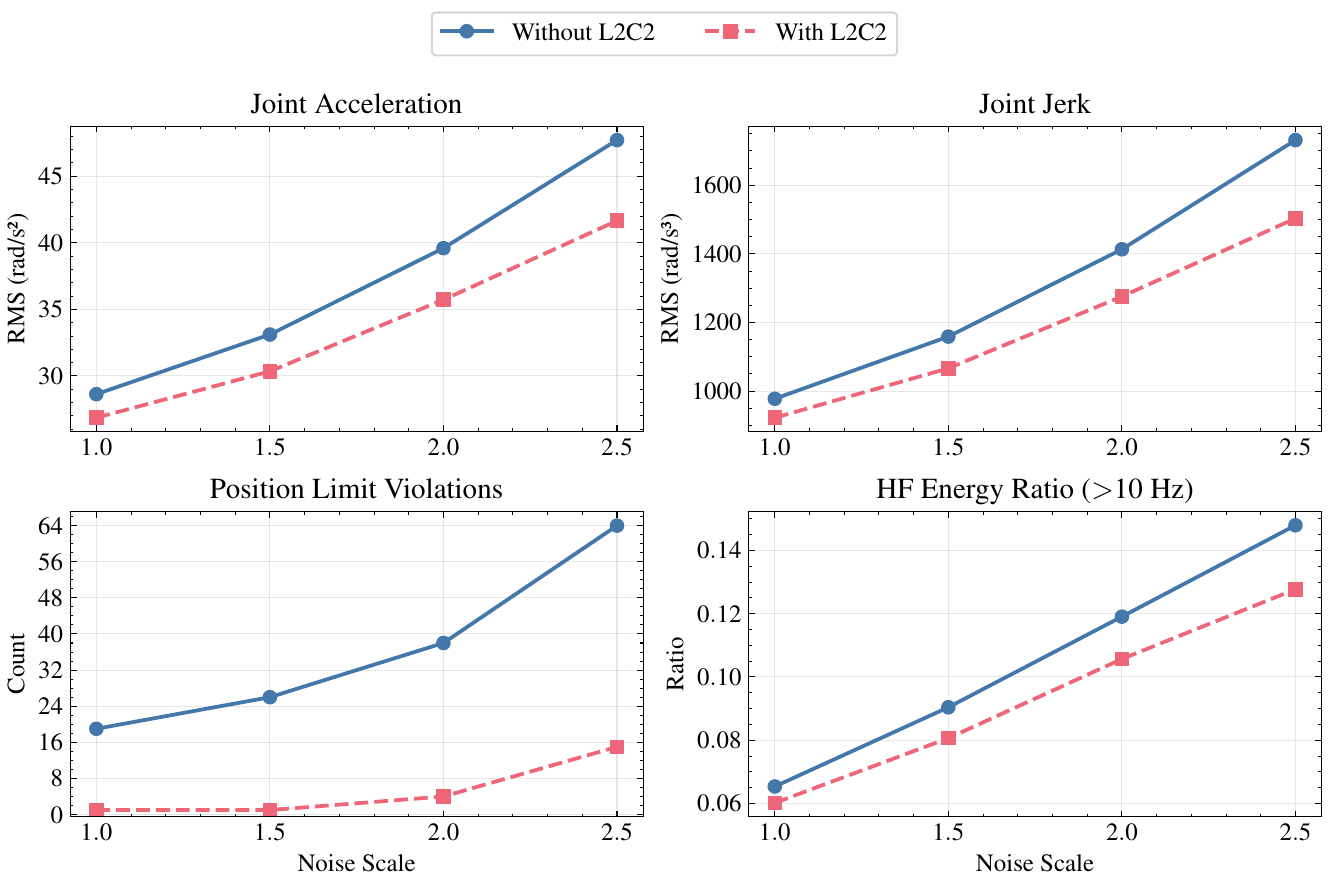}
    \caption{Effect of \ltctc{} on motion smoothness under increasing
    observation noise. Subplots: RMS joint acceleration, RMS joint jerk,
    position limit violations, and high-frequency energy ratio (fraction of
    joint-trajectory energy above 10\,Hz).}
    \label{fig:term_l2c2}
\end{figure}

We evaluate \ltctc{} on a mimic-style task using our MuJoCo evaluation pipeline ($15$ deterministic rollouts, $8s$ each). Both policies successfully transfer to real hardware for the same dancing motion; however, without \ltctc{} the actuators produce audible high-frequency oscillations. We quantify this via four deployment-critical metrics (\figref{fig:term_l2c2}): RMS joint acceleration, RMS joint jerk, joint limit violations, and high-frequency energy ratio. L2C2 consistently reduces all four metrics, with the benefit increasing along the noise level.

\subsubsection{Value-Bootstrapped Terminations}
\label{sec:ablation_termination}

\figref{fig:combined_ablations}(b) compares value-bootstrapped terminations
($\termsigma{=}5$) against a manually tuned termination penalty on \tone{}
stand-up (5 seeds). The bootstrapped variant converges to a higher
time-out ratio (fewer bad terminations) with substantially lower
seed variance (min/max shown), indicating more robust training. The
bootstrapped version requires no per-task penalty tuning; the same
$\termsigma$ works across all tasks.

\subsubsection{Virtual Harness}
\label{sec:ablation_harness}

\figref{fig:combined_ablations}(c) shows the effect of the virtual harness on
\gone{} height-controlled locomotion (5 seeds). With the harness (decayed over
the first 2k iterations), training recovers from the initial negative-reward
phase faster and converges to higher final reward. Without the harness, the
policy spends longer in unstable regimes and converges lower, with higher
variance across seeds.

\subsubsection{Symmetry Augmentation}
\label{sec:ablation_symmetry}

\begin{figure}[t]
    \centering
    \includegraphics[width=\linewidth]{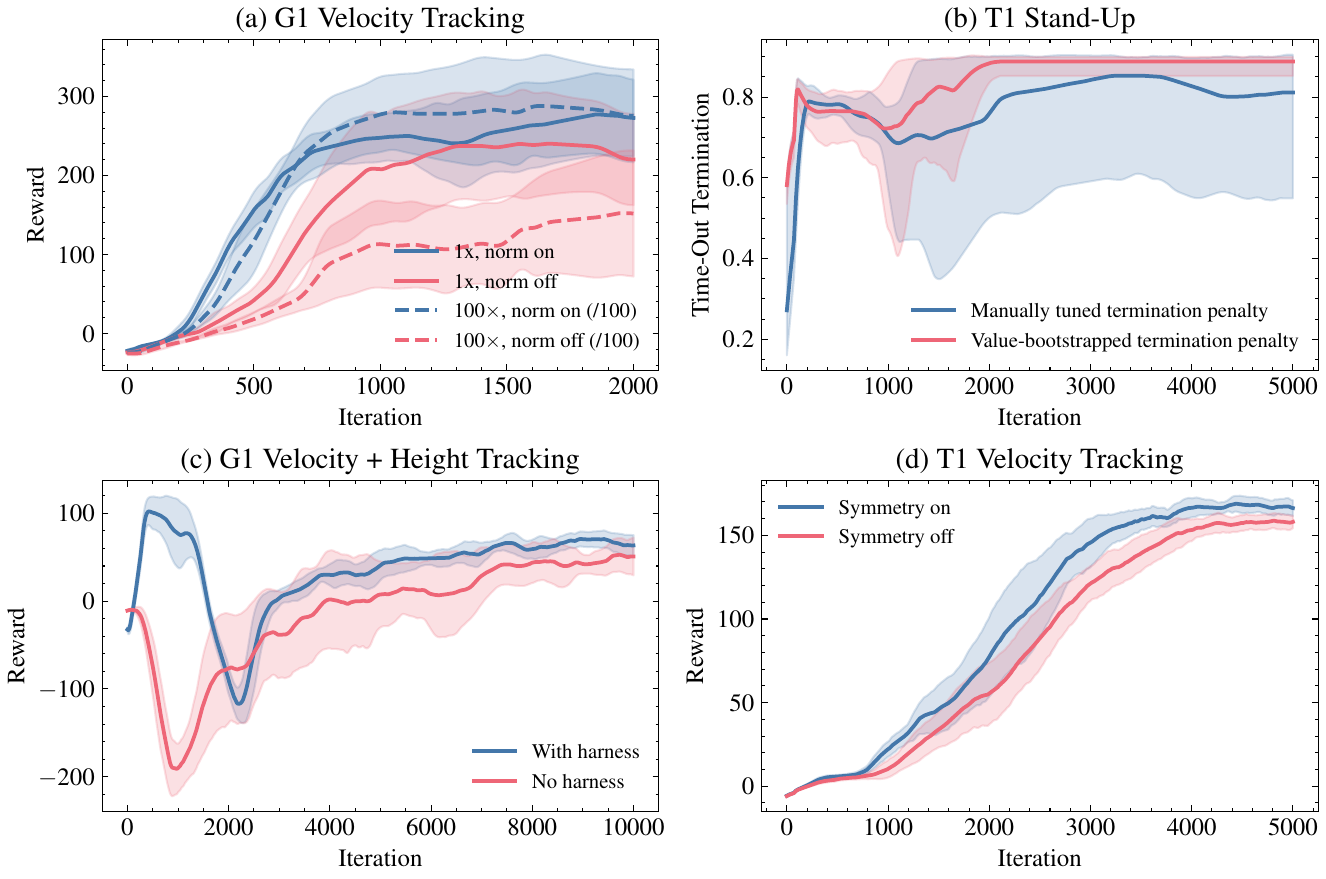}
    \caption{Ablation studies (5 seeds, shaded = $\pm1$std unless noted).
    (a)~Reward normalization; dashed curves scale all rewards by 100$\times$
    before training and rescale for comparability.
    (b)~Value-bootstrapped vs.\ tuned termination penalty
    (shaded = min/max).
    (c)~Virtual harness on height-controlled locomotion.
    (d)~Symmetry augmentation.}
    \label{fig:combined_ablations}
\end{figure}

\figref{fig:combined_ablations}(d) shows symmetry augmentation on \tone{}
velocity tracking. Mirroring observations and actions doubles effective data per batch and enforces symmetric gaits. The reward improvement is modest but consistent across all seeds; the primary benefit is behavioral symmetry, not captured by the reward curve.

\subsection{Sim-to-Real Transfer}
\label{sec:sim2real}

\begin{figure*}[t]
    \centering
    \includegraphics[width=\linewidth]{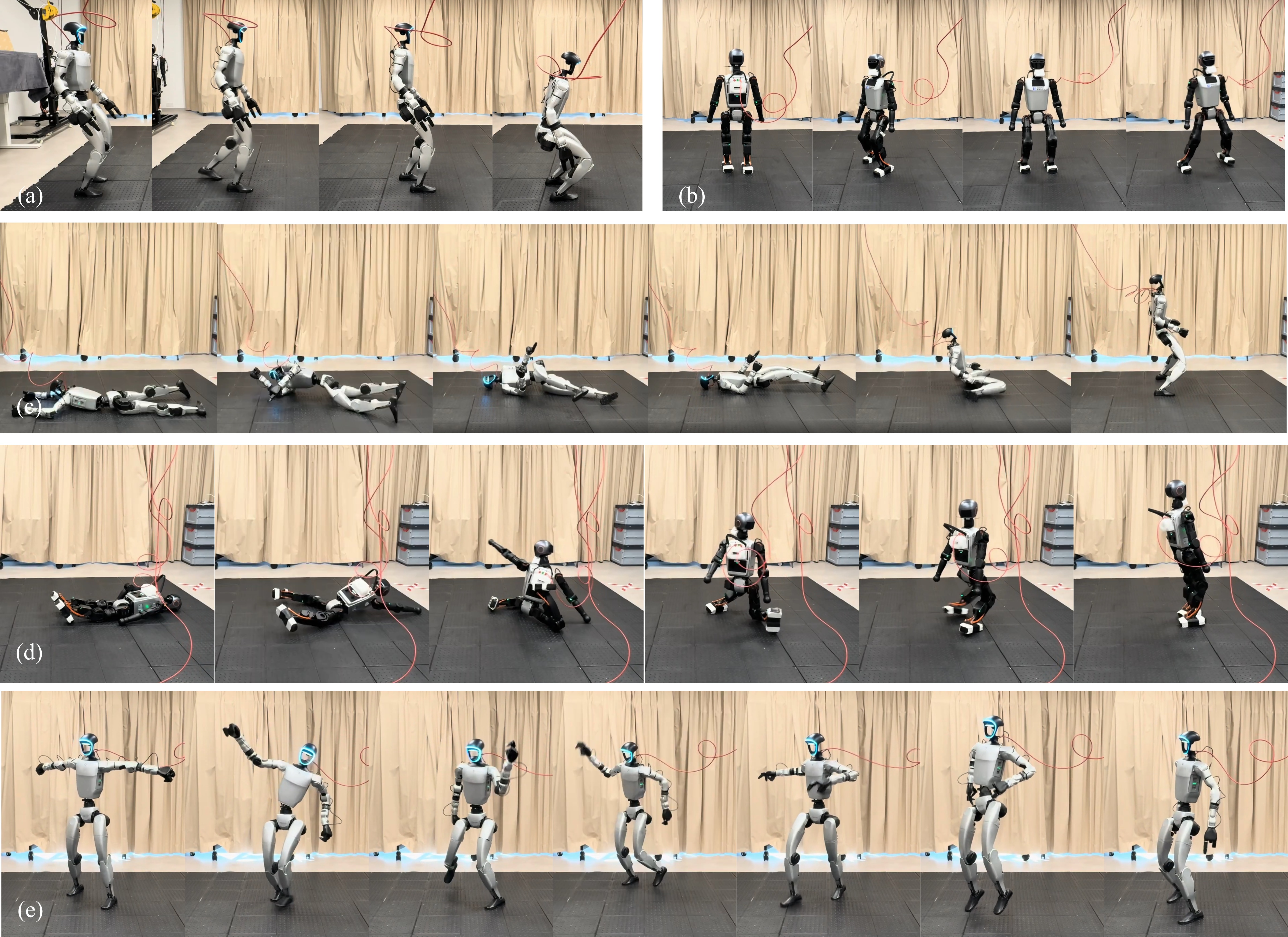}
    \caption{Sim-to-real transfer across five tasks on two robot platforms.
    (a)~Velocity and height controlled locomotion on \gone{}.
    (b)~Velocity controlled locomotion on \tone{}.
    (c)~Stand-up on \gone{}.
    (d)~Stand-up on \tone{}.
    (e)~Motion imitation (dancing) on \gone{}.}
    \label{fig:robot_motion}
\end{figure*}

All five tasks transfer to real hardware—with success defined by stable execution, absence of controller instability, and the completion of the designated task—while the loco-manipulation task is additionally validated in simulation via the VLA pipeline. \figref{fig:robot_motion} shows
stand-up on both robots and dancing on \gone{};
\figref{fig:vel_height_tracking} and \tabref{tab:tracking_rmse} report
sim-to-sim tracking metrics for height-controlled locomotion. The main
failure modes during development were sim-to-real gaps (actuator modeling,
contact dynamics) and overly aggressive policies producing motions that
real actuators cannot follow. Both were addressed by the domain
randomization and policy regularization described in
\secref{sec:system_design}. Since no external motion-capture system was available, sim-to-real transfer is validated qualitatively through hardware demonstrations; quantitative tracking metrics are reported via the \mujoco{} pipeline (\tabref{tab:tracking_rmse}). The sim-to-real pipeline will also be released in the near future separately. A supplementary video demonstrating all tasks on hardware is attached.

\subsection{Limitations \& Outlook}
\label{sec:limitations}

AGILE is currently validated on two humanoid platforms (Unitree G1 and Booster T1), and broader hardware validation remains future work. The framework builds on Isaac Lab, which simplifies integration with GPU-based simulation but introduces dependency on upstream APIs. In addition, the tasks studied here are primarily proprioceptive; perception-driven manipulation and more dynamic locomotion behaviors (e.g., running or stair climbing) are not yet included. Future work will expand robot and task coverage while continuing to validate new capabilities through real-world deployment.

\section{Acknowledgment}
\label{sec:acknowledge}

We thank Zhengyi Luo, Chenran Li and Xinghao Zhu for their discussion on modeling and training. We thank Rafael Wiltz, Sergey Grizan, Lotus Li, Tiffany Chen, David Chu and John Welsh for their discussion and feedback on tele-operation and data pipeline integration. We also thank H. Hawkeye King, Stephan Pleines and Vishal Kulkarni for their support on experiment and code release.


\bibliographystyle{IEEEtran}
\bibliography{references}

\appendix

\section{Appendix}

\subsection{Best Practices}
\label{app:best_practices}

Based on extensive experimentation across tasks and robots, we distill the
following practical guidelines:

\begin{enumerate}
  \item \textbf{Robot model validation.} Validate the USD model before any
        training. Spawn the robot on a flat ground plane in an interactive
        simulator, let it settle under gravity, and manually perturb
        it---the behavior should look physically plausible. Sweep all
        joints to their limits to verify correct signs and clamping,
        and use \agile{}'s joint debug GUI with symmetry mode to catch
        mirrored sign errors. A wrong joint axis wastes more GPU hours
        than any hyperparameter misconfiguration.

  \item \textbf{MDP validation.} Once the robot model is verified, validate
        the training environment before launching long runs. Confirm that a
        zero-action command produces a stable stand (catches incorrect
        default poses or action offsets), use the reward visualizer to
        verify that each reward term activates as expected, and check that
        observations have plausible values and ranges. These checks take
        minutes and prevent days of wasted training.

  \item \textbf{Reward composition.} Structure rewards into three groups:
        \emph{task} (what to achieve), \emph{style} (how it should look),
        and \emph{regularization} (what to avoid). As a rule of thumb,
        start simple: task rewards plus basic regularization to prevent
        unsafe behaviors, then incrementally add style and further
        regularization terms once the core task is solved.

  \item \textbf{Termination design.} If mean episode length collapses to
        near zero, the agent has become ``suicidal'': the expected return
        from continuing (accumulating negative rewards) is worse than
        terminating and resetting. The immediate fix is to increase the termination penalty
        (make it more negative), but this requires per-task tuning and
        can break as the value landscape shifts during training.
        Value-bootstrapped terminations (\secref{sec:ablation_termination})
        offer a more principled solution by making termination
        value-neutral and adding a fixed offset $\termsigma$, removing
        the need for per-task penalty tuning. However, because the
        value function's own predictions are bootstrapped at terminal
        states, inaccurate estimates can feed back into training
        targets. If value loss increases after enabling bootstrapped
        terminations, reduce $\termsigma$ or address the underlying
        reward signal first.
        Conversely, always terminate episodes when the robot enters
        unrecoverable states (e.g.\ fallen) to avoid wasting training
        compute on hopeless trajectories.

  \item \textbf{Curriculum design.} Two strategies: \emph{fading guidance}
        (start with assistance, remove it) and \emph{increasing difficulty}
        (start easy, ramp up penalties or terrain complexity). The harness
        force (\secref{sec:harness_force}) implements the former; terrain
        progression implements the latter.

  \item \textbf{Observation design.} Train teachers with privileged
        observations (terrain scans, contact forces, true velocities). Distill
        to students using only sensor-realistic inputs. Always add noise to
        policy observations matching expected sensor characteristics.

  \item \textbf{Training monitoring.} Track task metrics alongside reward
        curves: rising reward with stagnant task performance indicates reward
        hacking. Monitor value loss (should converge well below~$1.0$; if
        too high, scale down reward magnitudes or enable reward
        normalization to stabilize bootstrapping) and policy noise
        standard deviation (may increase initially during exploration but
        should decrease over the course of training; persistent growth
        signals that the entropy bonus dominates the task gradient). If noise keeps
        growing, first improve the reward function (cleaner advantages
        outweigh the entropy gradient), then reduce the entropy
        coefficient, or run a fraction of environments without domain
        randomization to guarantee clean gradient signal in every batch.
        Record periodic rollout videos---plots alone can be misleading.

  \item \textbf{Seed robustness.} Test across $\geq 5$ random seeds before
        drawing conclusions. A single lucky seed does not validate an MDP
        design. RL training is inherently sensitive to random seeds, and
        designing seed-robust environments is difficult; high variance
        across seeds often points to a fragile reward or curriculum
        design rather than bad luck.

  \item \textbf{Sim-to-real transfer.} Successful transfer rests on two
        pillars: \emph{robustness} via domain randomization and
        \emph{smoothness} via action regularization. For regularization,
        directly regularize the policy output via action norm,
        action rate (consecutive differences), and action acceleration
        (second-order differences) penalties; \ltctc{} further enforces
        smooth observation-to-action mappings. A policy that appears smooth in
        simulation may be relying on high simulated damping to mask
        aggressive actions---the policy \emph{itself} must output smooth
        commands. Domain randomization combined with these regularization
        terms were the most effective levers in our experience.
        History-based or recurrent policies can implicitly adapt to
        real dynamics at inference time, partially compensating for
        remaining sim-to-real gaps. If hardware behavior diverges
        significantly from simulation, the simulation is likely wrong:
        fix the simulation to match reality rather than compensating
        with reward shaping.
\end{enumerate}

\noindent
Finally, no single technique listed above works universally.
Reinforcement learning offers no magic trick that reliably helps across
all tasks and robots. \agile{} provides a toolbox of composable
modules, but each must be evaluated empirically for the problem at hand.

\subsection{Hyperparameter Tables}
\label{app:hyperparams}


\begin{table}[h]
\centering
\caption{\ppo{} hyperparameters. Task-specific overrides in parentheses.}
\label{tab:ppo_params}
\small
\begin{tabular}{@{}lc@{}}
\toprule
Parameter & Value \\
\midrule
Actor network          & $[256, 256, 128]$ (pick\&place: $[256, 128, 64]$) \\
Critic network         & $[512, 256, 128]$ (pick\&place: $[256, 128, 64]$) \\
Activation             & ELU \\
Learning rate          & $10^{-3}$ \\
Discount $\disc$       & $0.99$ (stand-up: $0.995$) \\
GAE $\lambda$          & $0.95$ \\
Clip ratio             & $0.2$ \\
Mini-batches           & 4 \\
Learning epochs        & 5 \\
Entropy coeff.         & $0.005$ (height: $0.01$, stand-up: $0.0025$) \\
Num.\ environments     & 4096 \\
Symmetry augmentation  & L-R mirror (except pick\&place) \\
\ltctc{}               & $\lactor{=}1.0$, $\lcritic{=}0.1$ \\
Reward normalizer      & $\decayf{=}0.999$, $\epsilon{=}0.01$ \\
\bottomrule
\end{tabular}
\end{table}

\end{document}